\pdfoutput=1

\documentclass[11pt]{article}

\usepackage[final]{ACL2023}
\usepackage{amsmath}
\usepackage{mathtools}
\usepackage{cleveref}
\usepackage{booktabs}
\usepackage{multirow}
\usepackage{adjustbox}
\usepackage{xcolor}

\usepackage{times}
\usepackage{latexsym}
\usepackage{bbm}
\usepackage{soul,color}
\usepackage{subcaption}
\usepackage{amsfonts}
\usepackage{enumitem}
\usepackage{hyperref}
\usepackage{makecell}

\usepackage[colorinlistoftodos]{todonotes}

\DeclareMathOperator*{\kNN}{\text{\textit{k}NN}}
\DeclareMathOperator*{\kNNLM}{\text{\textit{k}NN-LM}}
\DeclareMathOperator*{\LM}{LM}
\DeclareMathOperator*{\argmin}{arg\;min}

\DeclareMathOperator*{\GPT}{GPT-2}

\usepackage[T1]{fontenc}

\usepackage[utf8]{inputenc}

\usepackage{microtype}

\title{Adaptation Approaches for Nearest Neighbor Language Models}

 \author{Rishabh Bhardwaj$^1$\Thanks{Work done during an internship at AWS AI Labs.} \\\And George Polovets$^2$ \\ $^1$Singapore University of Technology and Design, Singapore \\ $^2$AWS AI Labs \\  \texttt{rishabhbhardwaj15@gmail.com} \\ \texttt{\{polovg, sunkaral\}@amazon.com} \\\And Monica Sunkara$^2$
}

\begin{document}
\maketitle
\begin{abstract}
Semi-parametric Nearest Neighbor Language Models ($\kNNLM$s) have produced impressive gains over purely parametric LMs, by leveraging large-scale neighborhood retrieval over external memory datastores. However, there has been little investigation into adapting such models for new domains. This work attempts to fill that gap and suggests the following approaches for adapting $\kNNLM$s --- 1) adapting the underlying LM (using Adapters), 2) expanding neighborhood retrieval over an additional adaptation datastore, and 3) adapting the weights (scores) of retrieved neighbors using a learned Rescorer module. We study each adaptation strategy separately, as well as the combined performance improvement through ablation experiments and an extensive set of evaluations run over seven adaptation domains. Our combined adaptation approach consistently outperforms purely parametric adaptation and zero-shot ($\kNNLM$) baselines that construct datastores from the adaptation data. On average, we see perplexity improvements of 17.1\% and 16\% for these respective baselines, across domains.

\end{abstract}

\section{Introduction}
Natural Language Processing (NLP) has observed large performance improvements with recent advancements in neural Language Models (LMs). These models have enabled learning rich, semantic text representations \cite{mikolov2010recurrent,bengio2000neural} that have facilitated a wide range of downstream language tasks \cite{radford2018improving, radford2019language}. For the task of next-word prediction, parametric LMs utilize the rich contextual text representations as input to a classifier (output layer), which produces a distribution over the possible next words.

\begin{figure}[ht]
\begin{adjustbox}{center}
\centering
    \includegraphics[width=0.5\textwidth]{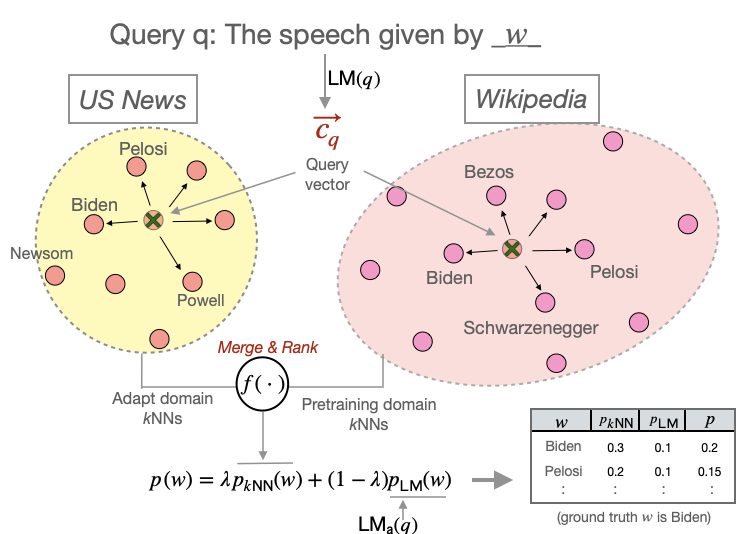}
\end{adjustbox}
\caption{An illustration of the proposed $\kNNLM$ adaptation approach. The current context is used as a query (q) for nearest-neighbor retrieval. The context is passed through the LM to obtain the query vector representation $\vec{c}_q$, which is then used to retrieve nearest neighbors from a large pretraining datastore and a smaller adaptation datastore (displayed in pink and yellow, respectively). The function $f(\cdot)$ represents merging of datastores (Merge), followed by rescoring (Rank) of the retrieved neighbors to obtain $p_{\kNN}$. The probability distribution over the candidate next words is computed by the mixture of probabilities $p_{\kNN}$ and $p_{\LM}$, where $p_{\LM}$ denotes probabilities obtained from domain-adapted LM.}
\label{fig:introduction}
\end{figure}

In contrast to parametric LMs, \textit{k}-Nearest Neighbor LMs ($\kNNLM$s) are semi-parametric models that maintain an external memory (i.e. \textbf{datastore}) \cite{khandelwal2019generalization}. This datastore is composed of key-value pairs, where the keys are contextual embeddings created from passing text data through an LM, and the values are the respective next-word labels. The datastore can be used to retrieve \textit{k}-nearest neighbors for the current context. The retrieved values induce a probability distribution over the next word, which is combined with the LM probabilities.

This mixture of probabilities has produced impressive gains over probabilities obtained from purely parametric LMs and has been shown to generate even larger improvements with the increase in the scale of the datastore \cite{khandelwal2019generalization, yogatama2021adaptive, he2021efficient}. While the dependency on a large-scale datastore is easy to satisfy when developing general-purpose pretrained models, it is challenging to develop effective $\kNNLM$s when it comes to specialized domains. This is due to the scarcity of domain-specific data, limiting the size of the corresponding datastore.

We posit that large, general-purpose datastores, referred to as the \textbf{pretraining datastore}, contain a significant amount of relevant information which can still be applied to specialized domains. This information can be leveraged through nearest-neighbor retrieval and should prove especially useful in situations where there is an insufficient amount of domain-specific data to generate an effective standalone datastore.

Unlike parametric neural architectures which can employ gradient-based finetuning for domain adaptation, it is less obvious how to adapt $\kNNLM$s primarily because of the non-parametric nature of datastores. One simple approach would be to reconstruct the datastore, using domain-adapted LM representations. However, this comes at the cost of incurring a large memory footprint for each adaptation domain. In this work, we instead choose to focus on adaptation strategies that are parameter and memory efficient.
Given the complementary nature of the parametric and non-parametric components in a $\kNNLM$, we pursue adaptation strategies separately for each component and analyze their impact on the $\kNNLM$ system's adaptation performance.
\begin{enumerate}
    \item Adaptation of the parametric LM: Given that we constrain ourselves to parameter-efficient adaptation techniques, we utilize Adapters \cite{houlsby2019parameter} for finetuning the parametric LM because of their competitive performance with full model finetuning \cite{Hou2022MetaLearningTD, Liu2022FewShotPF}. We also investigate the impact of adapting the parametric component on the quality of retrieved neighbors from the pretraining datastore.
    
    \item Adaptation of the non-parametric $\kNN$: As a memory-efficient alternative to reconstructing the pretraining datastore with domain-adapted representations, we formulate $\kNN$ adaptation as learning a domain-specific neighborhood scoring function (i.e. a \textbf{Rescorer}). This proposed Rescorer is trained to assign optimal weights to each retrieved neighbor for a given domain. We also consider expanding our neighborhood retrieval to include an additional datastore referred to as the \textbf{adaptation datastore}, created purely from the target domain. Relative to the pretraining datastore, the addition of the adaptation datastore further increases the memory footprint by an incremental amount.
\end{enumerate}
In line with previous works, we focus our experiments solely on the core Language Modeling task of next-word prediction (\citealp{khandelwal2019generalization}; \citealp{yogatama2021adaptive}). Results on seven adaptation domains ranging from science and books, to conversational text, demonstrate that our component-level strategies consistently improve over respective parametric and semi-parametric baselines, and produce even better results when combined together. Specifically, we find that adaptation of the parametric component increases recall of ground-truth labels found in the retrieved neighbors. We also confirm that the large-scale pretraining datastore contains relevant information for adaptation domains, via its performance edge over models that exclude it. Finally, we observe that expanding the nearest neighbor search to include elements from the adaptation datastore contributes to the best overall performing strategy. \Cref{fig:introduction} demonstrates the overall approach using Wikipedia and US News as example pretraining and adaptation domains, respectively.

\section{kNN-LMs} \label{sec:interpolation}
For a context $c_t$ defined by the sequence of words $(w_1, \ldots, w_{t-1})$, the causal language modeling task aims to model a probability distribution over the next word\footnote[1]{We use ``token'' and ``word'' interchangeably.} $w_t$. Let $p_{\LM}(w_t|c_t)$ and $p_{\kNN}(w_t|c_t)$ be the probability masses computed by the LM and $\kNN$ components, respectively. Details on how $p_{\kNN}(w_t|c_t)$ is computed and combined with $p_{\LM}(w_t|c_t)$ to produce the final $\kNNLM$ predictions, are outlined in the following sections.

\paragraph{Datastore creation:} Given a source domain training set $\mathcal{X}_s$, let $c_i= (w_1, \ldots, w_{t-1})$ be a sequence in $\mathcal{X}_s$. The datastore is defined as a set of $D_s$ tuples $\{(\vec{c}_i, w_i)\}_{i=1}^{D_s}$, where the key $\vec{c}_i\in{\mathbb{R}^{d_h}}$ denotes the contextual representation of $c_i$, produced by the LM and value $w_i$ denotes the next word label in the sequence.
\paragraph{\textit{k}-Nearest neighbor retrieval:} During inference, we obtain a query vector $\vec{c}_q\in{\mathbb{R}^{d_h}}$ for $\kNN$ retrieval by producing the contextual LM representation for the current sequence of tokens $c_q$. The neighborhood of $\vec{c}_q$ is constructed by retrieving its k nearest instances from the datastore. Let $\mathcal{D}(\cdot): \mathbb{R}^{2d_h} \to \mathbb{R}$ refer to the distance measure\footnote[2]{In practice large-scale datastores utilize approximate search methods for retrieval, detailed further in \hyperref[setup]{Section 4.1}.}. The \textit{k}-nearest neighbors of $\vec{c}_q$ can be obtained by:
\begin{align} \label{eq:nearestneighbor}
    \mathcal{K} &\coloneqq \underset{\text{k}}{\argmin}\: \{\mathcal{D}(\vec{c_q}, \vec{c_i})\}_{i\in[D_s]}
\end{align}
where k in the subscript denotes indices in $[D_s]$=$\{1,\ldots,D_s\}$ which corresponds to k smallest distances. The score (weight) $s_i$ of a neighbor key $\vec{c}_i$ is defined as:
\begin{align}
    s_i \coloneqq ||\vec{c_q}-\vec{c_i}||^2, i\in{\mathcal{K}}
\end{align}
Thus, the $\kNN$ probability of the next word can be obtained via:
\begin{equation} \label{eq:score}
    p_{\kNN}(w_t|c_t) \propto \sum_{i\in\mathcal{K}} \mathbbm{1}[{w_i{=}w_t}] \exp(-s_i) .
\end{equation}

\paragraph{Unifying $\kNN$ and LM:} The probability distribution of the $\kNNLM$ system can be obtained by interpolating the component probabilities
\begin{align} \label{eq:interpolation}
    \begin{split}
    p_{\kNNLM}(w_t|c_t) &= \\\lambda\;p_{\kNN}(w_t|c_t) &+ (1-\lambda)\;p_{\LM}(w_t|c_t)
    \end{split}
\end{align}
where $\lambda \in [0,1]$.

Since each probability distribution lies on a simplex spanning the token vocabulary, performing a convex combination of the two maintains a valid probability distribution.

\section{\textit{k}NN-LM Adaptation}
\subsection{Retrieval Quality Metrics} Beyond tracking LM perplexity improvement, we also introduce two simple metrics to measure the relevance and quality of retrieved neighborhoods. For neighborhood relevance, we define Recall as the fraction of times a ground-truth word is in the retrieved set of neighbors. For neighborhood quality, we denote Precision as the fraction of times the $\kNN$ assigns more probability to the ground truth token than the LM. We define:
\begin{align*}
     \text{Precision} &= \sum\limits_{t=1}^{N}\frac{ \mathbbm{1}[{p_{\LM}(w_{t}^*|c_t) < p_{\kNN}(w_{t}^*|c_t)}]}{N},\\
    \text{Recall} &=\sum\limits_{t=1}^{N}\frac{ \mathbbm{1}[{w_{t}^* \in \mathcal{K}_t}]}{N}.
\end{align*}
where $\mathcal{K}_t \coloneqq \{\vec{w_i}: {i\in [\mathcal{K}]}\}$, $w_{t}^*$ is the ground truth next word for the context $\vec{c_t}$, and $N$ is the total number of words in the dataset.
\subsection{Parametric LM Adaptation} \label{lm-adaptation}
We follow a parameter-efficient adaptation approach by keeping the pretrained LM fixed and learning Adapter modules, attached to all the model layers \cite{houlsby2019parameter}. Henceforth, we use \textbf{LM\textsubscript{a}} to denote the domain-adapted LM.

While Adapter-based models have shown performance on par with model fine-tuning on various adaptation tasks across domains \cite{pfeiffer2020adapterhub}, the impact of LM adaptation in a semi-parametric setting remains unclear. Given our constraint to keep the pretraining datastore keys static, updates to the parametric LM could create a mismatch between contextual representations used for querying and result in meaningless retrieved neighborhoods. We posit that Adapter-based LMs do not suffer from this because they preserve the metric space induced by the LM contextual encodings\footnote[3]{This is due to Adapters keeping the pretrained LM weight matrices frozen, thus preserving the coordinate space that is projected onto, when extracting contextual representations.}. Adapters tune representations in the original space such that they are more relevant to the adaptation domain.

\begin{quote}
\textit{Hypothesis-1} \label{sec:hyp1}: LM adaptation with Adapters not only assists the parametric models to perform better ($\downarrow$ perplexity), but also improves the quality of neighborhoods retrieved from the pretraining datastore ($\uparrow$ Recall).
\end{quote}
%


\subsection{\textit{k}NN Adaptation}
Given that we choose to keep the memory footprint of our adaptation approaches small (relative to the pretraining footprint), we fix the pretraining datastore representations (and thus the Recall of the retrieved neighborhoods) and instead focus on improving the Precision. This leads to our second hypothesis: 
\begin{quote}
\textit{Hypothesis-2} \label{sec:hyp2}: Using squared L2 distance between the query and neighbor key vectors is not the optimal neighbor scoring scheme. Instead, a more optimal scoring function can be learned for each domain.
\end{quote}
We propose learning a domain-optimized scoring function (a Rescorer) that learns to assign more weight to retrieved neighbors containing the ground-truth label.
%
%
We discuss the setup and architecture for the Rescorer in more detail subsequently.

\begin{figure}[ht]
\centering
    \includegraphics[width=0.48\textwidth]{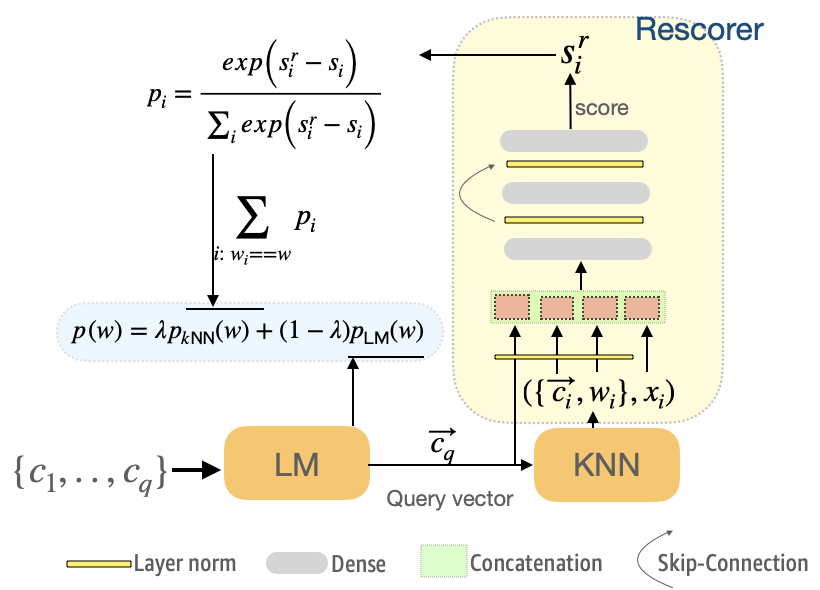}
    \caption{$\kNNLM$ Rescorer workflow. Neighbors extracted using the query contextual embedding are passed to a fully-connected Rescorer network. The scores output from the network are used to produce adapted nearest-neighbor probabilities $p_{\kNN}$, which are used in the final $p_w$ calculation.}
    \label{fig:rescorer}
\end{figure}

\paragraph{Rescorer Formulation:} Given a query vector $\vec{c_q}$ obtained from LM\textsubscript{a}, we retrieve a large set of neighbors $\mathcal{K}$. Each retrieved neighbor tuple $(\vec{c}_i, w_i) \in \mathcal{K}$ is passed through a neural module to obtain a domain-optimized score $s_i^r$. Let $f^r(\cdot): \mathbb{R}^{d_r} \to \mathbb{R}$ denote the Rescorer function. Its input is a set of three vectors: query $\vec{c_q}$, neighbor key vector $\vec{c_i}$, token embedding of the neighbor value $w_i$, as well as six features $\vec{x_i}= \{x_1, \ldots, x_6\}$ obtained from the pairwise dot products and pairwise euclidean distances between these three vectors\footnote[4]{We find that using these extra features produces the best quality Rescorer.}. The total input dimension is; $d_r$ = $3d_h + 6$ where $d_h$ is the dimension of the LM contextual representation. The final neighbor score $s_i'$ can be computed as\footnote[5]{We empirically observed that combining the learned and distance-based scores produces the best results.}:
\begin{align}
    s_i^r &= f^r([\vec{c_i}, \vec{c_q}, \vec{w_i}, \vec{x_i}])\\
    s_i' &= s_i^r - s_i \label{eq:score_new}
\end{align}
\paragraph{Rescorer Architecture:} We employ a three-layer fully-connected network for the Rescorer architecture. The input vectors are first layer-normalized and concatenated. They are then passed through two ReLU-activated dense layers with a skip connection $\mathbb{R}^{d_r} \to \mathbb{R}^{128} \to \mathbb{R}^{128}$ and a final dense (regression) layer $\mathbb{R}^{128} \to \mathbb{R}^1$ to generate the neighbor's score. The overall Rescorer workflow is shown in \Cref{fig:rescorer}.
\\
\paragraph{Rescorer Training:} We train the Rescorer to discriminate neighbors containing the ground truth as their values, by employing Contrastive Learning. We construct positive examples for retrieved neighbor tuples $(\vec{c_i}, w_i)$ if $w_i$ corresponds to the correct ground-truth next word, otherwise they are treated as negatives. We collect contextual embeddings for one million tokens from the adaptation domain training split\footnote[6]{If the training set has less than one million tokens, we utilize all of its tokens.} $\{w_1, \ldots, w_{\text{1M}}\}$ along with their nearest neighbors. Contrastive training examples are discarded if the ground-truth word is not found in the neighborhood values. From each neighborhood, the highest-scored (distance-based) positive neighbor is selected and 10 negative neighbors are randomly sampled. Contrastive Loss \cite{oord2018representation} is used to learn the Rescorer parameters and is defined as:
\begin{equation} \label{eq:rescorer_loss}
    \mathcal{L} = -\log\frac{\exp{(\frac{s_p^r}{\tau}})}{\exp{(\frac{s_p^r}{\tau}})+\sum\limits_{n} \exp{(\frac{s_n^r}{\tau}})}
\end{equation}
where $s_p^r$ and $s_n^r$ denote the Rescorer scores assigned to the positive and negative examples, respectively, and $\tau$ is a temperature hyperparameter.

\subsection{Merging $\kNN$s} 
While regenerating the pretraining datastore using an adapted LM (\Cref{lm-adaptation}) is generally a very memory-intensive procedure, creating a separate datastore purely from adaptation training data is expected to increase the memory footprint by a relatively small amount\footnote[7]{In our experimental setup, this amounts to 1-10\% relative increase in memory footprint}. With the availability of both pretraining and adaptation datastores, a natural extension is to utilize both during neighborhood retrieval. We extract the nearest neighbors independently from the pretraining datastore $\mathcal{K}_w$ and adaptation datastore $\mathcal{K}_a$ and merge them to create $\mathcal{K}_a \cup \mathcal{K}_w$. 
\\
\subsection{Adaptation of \textit{k}NN-LMs}
We summarize the overall adaptation strategy outlined in prior sections as follows: 
\begin{enumerate}
   \item Updating the parametric LM using lightweight Adapter modules.
   \item Merging the retrieved neighbors from the pretraining and adaptation datastores into a single neighborhood.
   \item Training a Rescorer with Contrastive Loss, to learn domain-optimal scores for retrieved neighbors.
\end{enumerate}

In the following results sections, we confirm the validity of \hyperref[sec:hyp1]{Hypothesis-1} and \hyperref[sec:hyp2]{Hypothesis-2}, as well as the efficacy of our simple neighborhood merging scheme through ablation studies. We also investigate the benefit of our collective adaptation strategy on modeling downstream domains.

\section{Experiments} \label{sec:exp_setup}
\subsection{Experimental Setup} \label{setup}

For all of our experiments, we utilize the off-the-shelf $\GPT$ \cite{radford2019language} model from Huggingface Transformers \cite{Wolf2019HuggingFacesTS}, as the pretrained LM. This model contains 117 million parameters with a vocabulary size of 50,257 word units, and directly matches the decoder-only configuration used in \citet{khandelwal2019generalization}. 
For the adaptation setting, Adapter modules are added to each layer of the pretrained GPT-2 resulting in 0.7\% extra parameters during finetuning. Training the Rescorer also amounts to learning an incremental 320K parameters, or roughly 0.3\% additional parameters relative to that of GPT-2. The Rescorer and Adapters are trained (separately) using AdamW optimizer with learning rates of 0.001 and 0.0001, respectively and a weight decay of 0.01. For sampling positive and negative examples during Rescorer training, we utilize a liberally sized neighborhood of size k=1000. Logging is performed every 200 iterations and early stopping is performed if there is no validation performance improvement for up to three logging steps.

The pretraining datastore is constructed from running GPT-2 on 1 billion tokens sampled from Wikipedia\footnote[8]{\cite{wikipedia}--\url{https://huggingface.co/datasets/wikipedia}} (i.e. $\mathcal{K}_w$) and any adaptation datastores are constructed from up to 100 million tokens taken from the training split of adaptation domains (i.e. $\mathcal{K}_a$). 
We select seven datasets across multiple domains to evaluate the performance of our adaptation strategies: XSum \cite{xsum} and XL-Sum \cite{xlsum} covering the news domain; SciQ \cite{sciq} and arXiv \cite{arxiv} for the science domain; BookSum \cite{booksum} for the literature domain, SAMSum \cite{samsum} for the conversational domain, and GovReport \cite{govreport}  for the government domain. For any summary-related datasets, we only utilize the original document text for our purposes and exclude summary ground-truth text. \Cref{tab:dstore_size} provides a breakdown of the resulting adaptation datastore sizes.

For nearest neighbor retrieval, we use FAISS - a library designed for fast similarity search in high dimensional space \cite{faiss}. Similar to \citet{khandelwal2019generalization}, we observe that L2-based FAISS search obtains better results than the inner-product, so we adopt this setting for our work as well. For all experiments, we perform hyperparameter search over k where k${\in}\{1,2,4,\ldots,512,1000\}$ and the $\kNN$ interpolation parameter $\lambda{\in}\{0.01, 0.02, 0.04, \ldots, 0.98\}$. 

\begin{table}[]
\centering
\resizebox{0.5\textwidth}{!}{
\begin{tabular}{|c|c|c|c|c|c|c|}
\hline
Xsum & SciQ & arXiv & BookSum & SAMSum & XLSum & GovSum \\ \hline
99.5 & 1.0 & 77.3 & 4.5 & 0.4 & 100.0 & 10.5 \\
\hline
\end{tabular}
}
\caption{Adaptation datastore size (in millions of entries).}
\label{tab:dstore_size}
\end{table}
\subsection{Models Used for Evaluation}
Because our work is the first to explore the intersection of LM adaptation with semi-parametric LMs, we use relevant baselines from both individual methodologies to track our modeling improvements. We provide the pretraining \textbf{(w)$\kNN$} and adaptation \textbf{(a)$\kNN$} neighborhood retrieval perplexities for reference\footnote[9]{\textbf{(w)$\kNN$} is obtained by putting $\lambda=0.9999$ in \Cref{eq:interpolation} to tackle cases where ground truth is not present in the retrieved neighborhood.}, to illustrate relevance of the pretraining domain to target domains and relationship between retrieval quality and datastore size. For the LM adaptation baseline, we compare against the performance of parametric finetuning with Adapters \textbf{LM$_a$}. For the semi-parametric LM baselines, we use two types of zero-shot evaluations of the $\kNNLM$. One applies zero-shot evaluation using the pretrained datastore \textbf{(w)$\kNNLM$} and the other evaluates using a datastore constructed out of the adaptation domain training data \textbf{(a)$\kNNLM$}. The latter strategy, also presented in \citet{khandelwal2019generalization}, to the best of our knowledge, is the only other work that utilizes adaptation data with $\kNNLM$s.

Beyond these models, we perform extensive experimentation with different combinations of datastores to use for retrieval (Wikipedia - \textbf{(w)}, Adaptation training split - \textbf{(a)}, Both - \textbf{(w+a)}), types of parametric LMs (Pretrained LM - \textbf{LM}, Adapted LM - \textbf{LM$_a$}), and usage of Rescorers (Rescorer used - \textbf{$\kNN_r$}, No Rescorer used - \textbf{$\kNN$}). These combinations provide precise ablations of our adaptation component improvements and their interactions with one another. 

\section{Results and Discussions}
\subsection{Hypothesis Pilot Studies}
We first motivate our larger experimental effort with pilot studies that test out \hyperref[sec:hyp1]{Hypothesis-1}, \hyperref[sec:hyp2]{Hypothesis-2}, and the neighborhood merging strategy. These studies are run on a subset of 100K test tokens taken from each adaptation domain. In our initial pilot experiments, we find that using k=1000 neighbors produces the best results. Any adaptation strategy requiring gradient-based optimization is performed on the respective adaptation training splits.

\paragraph{Evaluating Hypothesis-1:} To test this hypothesis, we measure the impact of LM adaptation on retrieval quality from the pretraining , by observing changes to the $\kNN$'s Recall value. \Cref{tab:lm_adaptation} demonstrates that adaptation of the parametric LM (LM$_a$) improves not only perplexity, but also retrieval Recall (retrieved neighbors using LM$_a$ are denoted by \textit{k}NN*, while neighbors retrieved with the pretrained LM are denoted by \textit{k}NN). This appears to support our hypothesis that techniques like Adapters, which preserve the LM representation space, can also benefit the retrieval component.

\begin{table}
\centering
\resizebox{0.48\textwidth}{!}{
\begin{tabular}{@{}lcccccc@{}}
\toprule
Domain  & \multicolumn{4}{c}{Perplexity ($\downarrow$)}                                & \multicolumn{2}{c}{Recall ($\uparrow$)} \\ \midrule
        & LM    & LM\textsubscript{a}  & \textit{k}NN & \multicolumn{1}{c|}{\textit{k}NN*}   & \textit{k}NN      & \textit{k}NN*       \\
XSum    & 22.45 & \textbf{18.95} & 83.95        & \multicolumn{1}{c|}{\textbf{74.67}}  & 88.72             & \textbf{89.29}      \\
SciQ    & 22.15 & \textbf{16.10} & 46.86        & \multicolumn{1}{c|}{\textbf{38.64}}  & 92.53             & \textbf{93.26}      \\
arXiv   & 56.83 & \textbf{24.97} & 513.44       & \multicolumn{1}{c|}{\textbf{270.11}} & 77.54             & \textbf{79.89}      \\
BookSum & 21.15 & \textbf{20.45} & 64.92        & \multicolumn{1}{c|}{\textbf{62.15}}  & 90.14             & \textbf{90.34}      \\
SAMSum  & 46.86 & \textbf{32.25} & 298.08       & \multicolumn{1}{c|}{\textbf{228.99}} & 96.36             & \textbf{96.64}      \\
XL-Sum   & 24.87 & \textbf{21.84} & 100.92       & \multicolumn{1}{c|}{\textbf{89.65}}  & 87.98             & \textbf{88.60}      \\
GovReport  & 19.31 & \textbf{14.72} & 83.62        & \multicolumn{1}{c|}{\textbf{66.91}}  & 88.55             & \textbf{89.47}      \\ \bottomrule
\end{tabular}
}
\caption{Hypothesis -1 pilot study. Adapted LM representations improve both perplexity and retrieval Recall.}
\label{tab:lm_adaptation}
\end{table}

\paragraph{Evaluating Hypothesis-2:} To test whether Rescorer-generated scores improve over purely distance-based scores, we contrast the resulting Precision of both types of scoring methods. 
\Cref{tab:knn_adaptation} shows that the domain-adapted scores produced by the Rescorer yield significantly higher neighborhood Precision on average than those using purely L2-based scoring. This applies for neighbors retrieved from the pretraining datastore (w)$\kNN_r$, as well as from datastores constructed from adaptation domain samples (a)$\kNN_r$. This suggests that the Rescorer can act as a general-purpose improvement over the standard $\kNNLM$ setup, regardless of whether neighbors are retrieved from in-domain or out-of-domain datastores. The improvement in Precision also confirms the efficacy of Contrastive Learning in producing a Rescorer that can discriminate between neighbors containing the ground-truth token from those that don't.

\begin{table}[]
\centering
\resizebox{0.5\textwidth}{!}{
\begin{tabular}{llccc}
\hline
        & \multicolumn{4}{c}{Precision ($\uparrow$)}                                                                    \\ \hline
Domain  & (w)$\textit{k}\text{NN}$ & (w)$\textit{k}\text{NN}_r$ & (a)$\textit{k}\text{NN}$ & (a)$\textit{k}\text{NN}_r$ \\
XSum    & 29.6                   & \textbf{44.9}            & 45.9                   & \textbf{59.8}            \\
SciQ    & 33.9                   & \textbf{48.2}            & 45.8                   & \textbf{53.0}            \\
arXiv   & 25.6                   & \textbf{38.2}            & 52.8                   & \textbf{65.4}            \\
BookSum & 33.1                   & \textbf{54.7}            & 33.7                   & \textbf{50.1}            \\
SAMSum  & 25.9                   & \textbf{27.7}            & 37.0                   & \textbf{38.6}            \\
XL-Sum   & 29.9                   & \textbf{46.7}            & 43.9                   & \textbf{58.5}            \\
GovReport  & 25.7                   & \textbf{42.6}            & 43.7                   & \textbf{55.9}            \\ \hline
\end{tabular}
}
\caption{Hypothesis-2 pilot study. Applying the Rescorer leads to Precision improvements in neighbors retrieved from both pretraining and adaptation datastores.}
\label{tab:knn_adaptation}
\end{table}

\begin{table}[]
\centering
\resizebox{0.45\textwidth}{!}{
\begin{tabular}{@{}lccc@{}}
\toprule
\multicolumn{1}{c}{} & \multicolumn{3}{c}{Recall ($\uparrow$)}                                                              \\ \midrule
Domain               & \multicolumn{1}{l}{(w)$\textit{k}\text{NN}$} & (a)$\textit{k}\text{NN}$ & (w+a)$\textit{k}\text{NN}$ \\
XSum                 & 89.3                                         & 92.7                     & \textbf{93.1}              \\
SciQ                 & 92.5                                         & 91.5                     & \textbf{94.7}              \\
arXiv                & 79.9                                         & 91.9                     & \textbf{92.1}              \\
BookSum              & 90.3                                         & 86.8                     & \textbf{91.5}              \\
SAMSum               & 84.3                                         & 85.7                     & \textbf{88.9}              \\
XL-Sum                & 88.6                                         & 92.1                     & \textbf{92.5}              \\
GovReport               & 89.5                                         & 92.5                     & \textbf{93.6}              \\ \bottomrule
\end{tabular}
}
\caption{Recall improvement from merging \textit{k}NNs. Merged neighborhoods consistently obtain higher Recall than those obtained from the individual datastores.}
\label{tab:merge_recall}
\end{table}

\paragraph{Effectiveness of Neighborhood Merging} To test the effectiveness of the simple neighborhood merging strategy, we contrast the Recall of merged neighborhoods to those of standalone neighborhoods from each individual datastore. In this  study, we keep the total number of retrieved neighbors fixed and empirically find that retrieving 500 nearest neighbors from each datastore in the merging strategy works best. The results of this study (\Cref{tab:merge_recall}) show that the combined set of neighbors $\mathcal{K}_a \cup \mathcal{K}_w$ has a better Recall value than either individual neighborhood. Due to this observed Recall improvement, we use this simple merging technique in our overall adaptation strategy. When training a Rescorer on these merged neighborhoods, we pass an additional binary input feature to inform the model on which datastore a particular neighbor comes from.

\begin{table*}[]
\centering
\resizebox{1.00\textwidth}{!}{
\begin{tabular}{lccccccccccccc}
\hline
\multicolumn{1}{c}{}                                                                                                              &                                                                 & \multicolumn{5}{c}{\textbf{Configuration}}                                                             & \multicolumn{7}{c}{\textbf{Perplexity ($\downarrow$)}}                                                                   \\ \hline
\multicolumn{1}{c}{}                                                                                                              & \textbf{Setting}                                                & \textbf{LM}  & \textbf{LMa} & \textbf{(w)KNN} & \textbf{(a)KNN} & \multicolumn{1}{c|}{\textbf{rescore}} & \textbf{XSum}  & \textbf{SciQ}  & \textbf{arXiv} & \textbf{BookSum} & \textbf{SAMSum} & \textbf{XL-Sum} & \textbf{GovReport} \\ \hline
\multicolumn{1}{l|}{\multirow{4}{*}{Baseline}}                                                                                    & \multicolumn{1}{c|}{$\text{LM}$ (only)}                         & $\checkmark$ &              &                 &                 & \multicolumn{1}{c|}{}                & 22.45          & 22.15          & 56.83          & 21.15            & 46.86           & 24.87          & 19.32           \\
\multicolumn{1}{l|}{}                                                                                                             & \multicolumn{1}{c|}{$\text{LM}_a$ (only)}                       &              & $\checkmark$ &                 &                 & \multicolumn{1}{c|}{}                & 18.95          & 16.09          & 24.97          & 20.45            & 32.26           & 21.84          & 14.72           \\
\multicolumn{1}{l|}{}                                                                                                             & \multicolumn{1}{c|}{(w)\textit{k}$\text{NN}$ (only)}            &              &              & $\checkmark$    &                 & \multicolumn{1}{c|}{}                & 83.96          & 46.86          & 513.45         & 64.92            & 298.08          & 100.92         & 83.62           \\
\multicolumn{1}{l|}{}                                                                                                             & \multicolumn{1}{c|}{(a)\textit{k}$\text{NN}$ (only)}            &              &              &                 & $\checkmark$    & \multicolumn{1}{c|}{}                & 38.38          & 57.82          & 87.58          & 109.87           & 229.89          & 47.75          & 40.39           \\ \hline
\multicolumn{1}{l|}{\multirow{2}{*}{\begin{tabular}[c]{@{}l@{}}Baseline\\ \shortcite{khandelwal2019generalization}\end{tabular}}} & \multicolumn{1}{c|}{(w)\textit{k}$\text{NN}$-$\text{LM}$}       & $\checkmark$ &              & $\checkmark$    &                 & \multicolumn{1}{c|}{}                & 21.64          & 19.19          & 53.03          & 20.50            & 46.27           & 24.03          & 18.99           \\
\multicolumn{1}{l|}{}                                                                                                             & \multicolumn{1}{c|}{(a)\textit{k}$\text{NN}$-$\text{LM}$}   & $\checkmark$ &              &                   &  $\checkmark$    & \multicolumn{1}{c|}{}                & 17.01          & 14.71          & 24.38          & 20.60            & 39.99           & 19.39          & 14.87           \\ \hline
\multicolumn{1}{l|}{\multirow{6}{*}{Ours}}                                                                                        & \multicolumn{1}{c|}{(w)\textit{k}$\text{NN}$-$\text{LM}_a$}     &              & $\checkmark$ & $\checkmark$    &                 & \multicolumn{1}{c|}{}                & 18.42          & 14.62          & 24.42          & 19.72            & 31.94           & 21.22          & 14.47           \\
\multicolumn{1}{l|}{}                                                                                                             & \multicolumn{1}{c|}{(w)\textit{k}$\text{NN}_r$-$\text{LM}$}     & $\checkmark$ &              & $\checkmark$    &                 & \multicolumn{1}{c|}{$\checkmark$}    & 21.32          & 18.5           & 51.89          & 20.09            & 46.20           & 23.68          & 18.81           \\
\multicolumn{1}{l|}{}                                                                                                             & \multicolumn{1}{c|}{(w)\textit{k}$\text{NN}_r$-$\text{LM}_a$}   & $\checkmark$ & $\checkmark$ & $\checkmark$    &                 & \multicolumn{1}{c|}{$\checkmark$}    & 18.23          & 14.22          & 24.19          & 19.35            & 31.92           & 20.98          & 14.36           \\
\multicolumn{1}{l|}{}                                                                                                             & \multicolumn{1}{c|}{(a)\textit{k}$\text{NN}$-$\text{LM}_a$}     &              & $\checkmark$ &                 & $\checkmark$    & \multicolumn{1}{c|}{}                & 15.30          & 12.88          & 17.81          & 20.22            & 31.48           & 18.12          & 13.08           \\
\multicolumn{1}{l|}{}                                                                                                             & \multicolumn{1}{c|}{(a)\textit{k}$\text{NN}_r$-$\text{LM}_a$}   &              & $\checkmark$ &                 & $\checkmark$    & \multicolumn{1}{c|}{$\checkmark$}    & 14.85          & 12.72          & 17.49          & 20.09            & 31.47           & 17.72          & 12.87           \\
\multicolumn{1}{l|}{}                                                                                                             & \multicolumn{1}{c|}{(w+a)\textit{k}$\text{NN}$-$\text{LM}_a$}   &              & $\checkmark$ & $\checkmark$    & $\checkmark$    & \multicolumn{1}{c|}{}                & 15.20          & 12.15          & 17.85          & 19.72            & 31.20           & 17.99          & 13.01           \\
\multicolumn{1}{l|}{}                                                                                                             & \multicolumn{1}{c|}{(w+a)\textit{k}$\text{NN}_r$-$\text{LM}_a$} &              & $\checkmark$ & $\checkmark$    & $\checkmark$    & \multicolumn{1}{c|}{$\checkmark$}    & \textbf{14.71} & \textbf{11.95} & \textbf{17.47} & \textbf{19.42}   & \textbf{31.18}  & \textbf{17.53} & \textbf{12.79}  \\ \hline
\end{tabular}
}
\caption{Performance of different $\kNNLM$ configurations. (w)\textit{k}NN, (a)\textit{k}NN, and (a+w)\textit{k}NN denote neighborhood search from pretraining datastore, adaptation datastore, and equal contribution from both, respectively. LM and LM\textsubscript{a} denote standard $\GPT$ and domain-adapted $\GPT$.}
\label{tab:knn_merge_adaptation}
\end{table*}

\subsection{Domain Adaptation Evaluations}
\Cref{tab:knn_merge_adaptation} compares the perplexities of the various models evaluated on the seven adaptation test sets. First we note that while the adapted LM yields the expected perplexity reductions over the pretrained LM (LM$_a$ < LM), we observe that zero-shot evaluation of the pretrained $\kNNLM$ also performs better than the pretrained LM ((w)$\kNNLM$ < LM). This continues to confirm the capacity of the pretraining datastore to retrieve relevant neighbors for downstream domains. We also find that in a majority of the cases, zero-shot evaluation of a $\kNNLM$ constructed over the adaptation datastore, outperforms parametric adaptation ((a)$\kNNLM$ < LM$_a$). This corroborates the finding from \citet{khandelwal2019generalization}, where utilizing data for neighborhood retrieval can outperform using it for LM training.

The results further support our \hyperref[sec:hyp1]{Hypothesis-1}, namely that parametric LM adaptation improvement is compounded when used in the $\kNNLM$ setting (e.g. $(w)\textit{k}$\text{NN}$_r$ < LM$_a$ < (w)$\kNNLM$$_a$). They also add support for \hyperref[sec:hyp2]{Hypothesis-2} where the Rescorer acts as a general-purpose improvement to $\kNNLM$ models (by noting that $\kNN_r$-based models outperform respective $\kNN$-based models). We observe that merging neighborhoods from both datastores also provides some small perplexity gains. Overall, our combined adaptation approach (last row of \Cref{tab:knn_merge_adaptation}) produces an average of 17.1\% and 16\% perplexity improvement over the parametric adaptation \textbf{LM$_a$} and semi-parametric baselines \textbf{(a)$\kNNLM$} respectively.

\begin{figure}[ht]
\centering
    \includegraphics[width=0.48\textwidth]{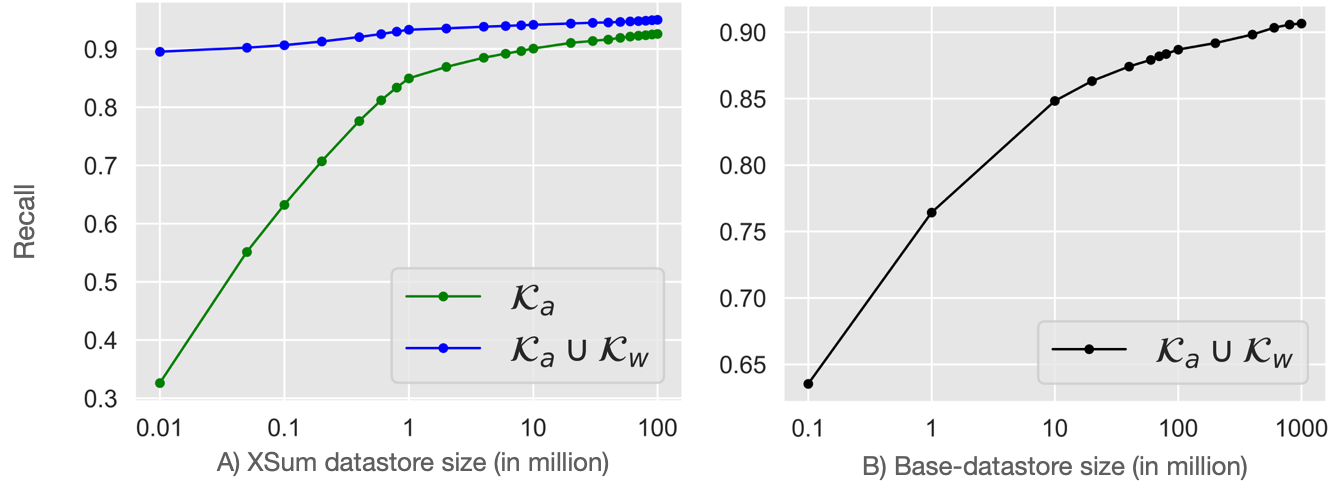}
    \caption{a) Change in Recall of merged neighborhoods compared against the size of the adaptation datastore; b) change in Recall of merged neighborhoods compared against the size of the pretraining datastore.
    }
    \label{fig: recall}
\end{figure}

\paragraph{Pretraining-datastore under low-resource adaptation.} We analyze the impact on Recall when combining neighbors from the pretraining and adaptation datastores in a low-resource adaptation setting (which is a common scenario). We utilize the Xsum dataset (containing nearly 100M training tokens), to analyze the impact of merging retrieved neighborhoods for different sizes of the adaptation datastore. In \Cref{fig: recall}-a), we observe that the Recall of retrieved neighbors significantly decreases as the adaptation datastore size decreases (green, $\mathcal{K}_a$). However, the merged neighborhood Recall enjoys a relatively flat curve (blue, $\mathcal{K}_a \cup$ $\mathcal{K}_w$). This suggests that the pretraining datastore acts as an important buffer in maintaining high-quality neighborhoods for low-resource adaptation scenarios. A complementary study to consider for the low-resource setting is the impact of the size of the pretraining datastore on the merged retrieval Recall. In this set of experiments, we fix the size of the adaptation datastore to be 100K. From \Cref{fig: recall}-b), we observe that Recall monotonically increases with the size of the pretraining datastore and may continue to improve even after the pretraining datastore exceeds 1 billion tokens. Thus, scaling the pretraining datastore can lead to improved retrieval quality on downstream domains.
\paragraph{Which LM representations are better for datastore construction?} An important question to consider, is which representations from $\GPT$ are most useful in constructing the datastore. To investigate this, we experiment with using different layers from GPT-2 in constructing a Wikipedia-based datastore. To increase the throughput experimentation, we use a smaller-sized datastore of size 10 million. We consider the output of the penultimate Transformer block as well as the following layers from the last Transformer block in our analysis: first layer norm (LN1), output of Multi-Headed Attention (MHA), second layer norm (LN2), output of final feed-forward layer (FFN). Thus, each datastore differs only in its key vector representations $\vec{c}_q$ for a given context $c_q$. The $\kNNLM$ probability is computed as per \Cref{eq:interpolation} where k is set to $1000$ and $\lambda$ is a hyperparameter tuned via grid search in $\lambda{\in}\{0.01, 0.02, 0.04, \ldots, 0.98\}$. Evaluation is performed on 100K test tokens obtained from unseen Wikipedia documents.

\begin{figure}[ht]
\centering
    \includegraphics[width=0.45\textwidth]{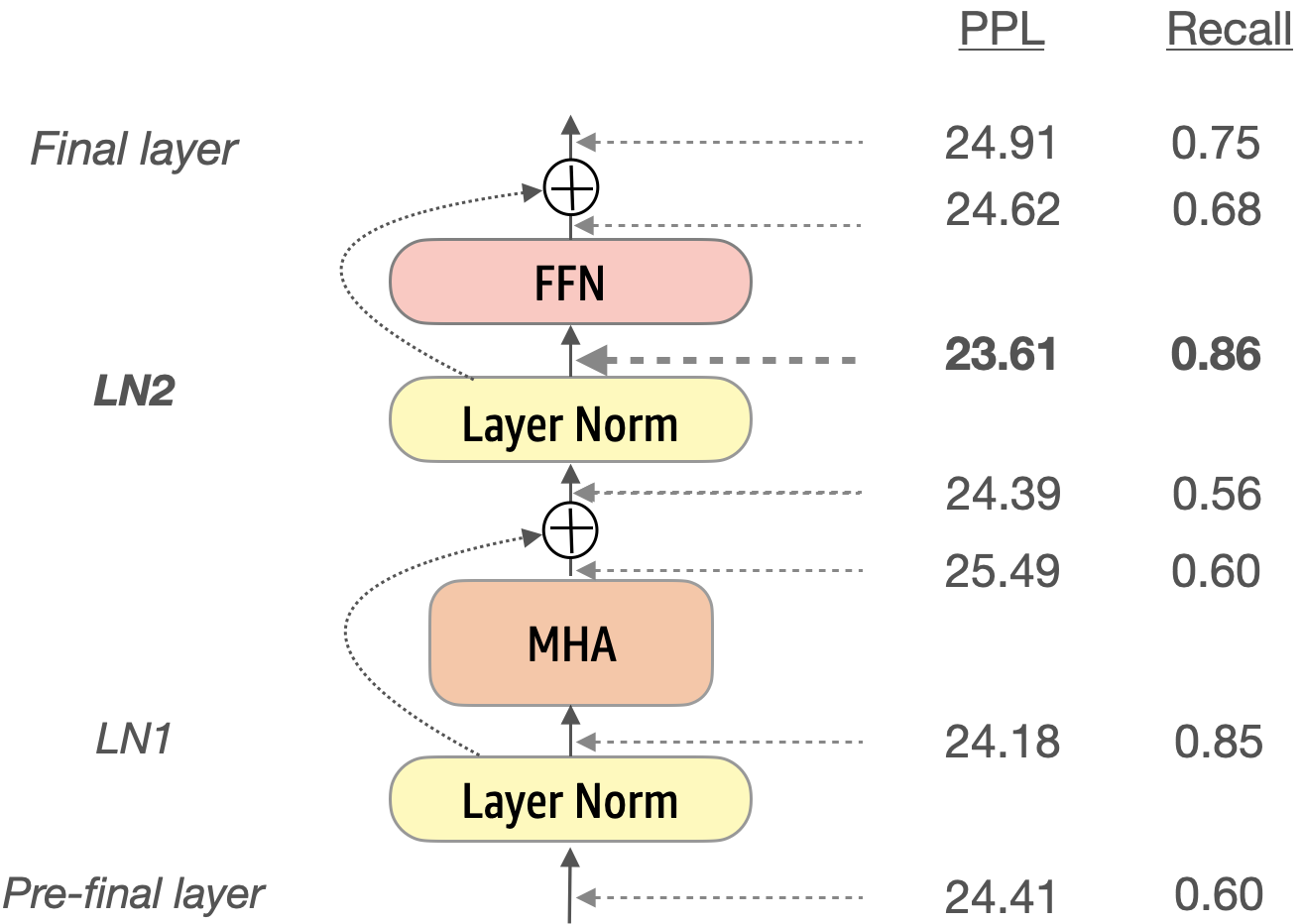}
    \caption{The figure shows the internals of $\GPT$ final layer. Perplexity (PPL) and Recall score of different context vector ($\vec{c}_q$) candidate representations for datastore construction.}
    \label{fig:pilot}
\end{figure}

As shown in \Cref{fig:pilot}, we observe that using the output of the LN2 layer creates the best representation space for the datastore keys and produces the best test perplexity of 23.61 and highest Recall of 0.86. We also observe that the best $\lambda$ returned for an LN2-based $\kNNLM$ is 0.1, which is the highest among context representation candidates considered.

\paragraph{Computational cost.}
We compare our computational overhead with respect to the  standard $\kNNLM$ proposed by \citet{khandelwal2019generalization}. During inference, an Adapter increases the inference time of $\GPT$ by about 1.2 milliseconds per token. The Rescorer takes about 60 milliseconds per token to score 1000 neighbors. We run the parametric model on a single GPU\footnote[10]{Tesla V100-SXM2-16GB} $\kNN$ and the Rescorer on CPU.

\section{Related work}
Our proposed work investigates the intersection of techniques used for parametric Language Model adaptation with semi-parametric systems ($\kNNLM$s). Therefore we discuss the related works in each of these areas and contrast our respective contributions.
\paragraph{Parametric LM Adaptation}
Popularization of Large-Scale Pretrained Language Models (PLMs) has necessitated research into parameter-efficient adaptation methods, to avoid maintaining large models for each domain. Many parameter-efficient methods keep the pretrained LM parameters frozen and learn additional layers during adaptation (\citealp{houlsby2019parameter}; \citealp{BenZaken2022BitFitSP}), or modify the parameters of existing layers (\citealp{Hu2022LoRALA}; \citealp{Hou2022MetaLearningTD}). This work explores how applying such techniques (namely Adapters) can improve the semi-parametric LM adaptation performance.
\paragraph{Semi-Parametric KNN-LMs}
Previous works have motivated that scaling the datastore for large-scale retrieval acts as a complimentary path to scaling data used for LM training (\citealp{khandelwal2019generalization}; \citealp{Borgeaud2022ImprovingLM}; \citealp{Khandelwal2021NearestNM}). However, adaptation approaches of these semi-parametric systems beyond zero-shot evaluation (\citealp{khandelwal2019generalization}; \citealp{Khandelwal2021NearestNM}) have not been explored up until this work.

To improve the quality of retrieval-enhanced methods, neighborhood Rescorer techniques have been employed for other domains such as Q\&A \citep{Glass2022Re2GRR} and information retrieval \citep{Nogueira2019PassageRW}. In contrast, this work explores applications of Rescorer techniques for the Language Modeling task and considers them for lightweight adaptation of semi-parametric LMs.

\section{Conclusion}
We proposed a multi-pronged strategy for adapting $\kNNLM$ systems. Through our studies, we demonstrated that a general-purpose pretraining datastore contains relevant information, which can be utilized for downstream domains. We showed that parametric and non-parametric adaptation methods complement each other and that using the complete semi-parametric adaptation strategy outperforms adapting just one of the $\kNNLM$ components. Our methods could further be extended by noting that the Recall of retrieved neighborhoods is often imperfect. Thus, a gate could be learned to predict whether $\kNN$ retrieval should be triggered. While our study focused on the Language Modeling task, our approach could be applied towards other NLP tasks such as text generation and translation.

\section{Acknowledgement}
The authors express their gratitude to Kyu Han and Shiva Sundaram for their continuous support throughout this work. They are also appreciative of Omid Sadjadi, Sundararajan Srinivasan, and Zejiang Hou for providing valuable feedback on the preliminary draft.



\newpage
\bibliography{main}
\bibliographystyle{acl_natbib}

\end{document}